\documentclass[10pt,twocolumn,letterpaper]{article}

\usepackage[pagenumbers]{iccv} % To force page numbers, e.g. for an arXiv version
\usepackage{array}
\usepackage{multirow}
\usepackage{makecell}

\definecolor{iccvblue}{rgb}{0.21,0.49,0.74}
\usepackage[pagebackref,breaklinks,colorlinks,allcolors=iccvblue]{hyperref}

\def\paperID{*****} % *** Enter the Paper ID here
\def\confName{ICCV}
\def\confYear{2025}

\title{MotivNet: Evolving Meta-Sapiens into an Emotionally Intelligent Foundation Model}

\author{Rahul Medicharla\\
Photogrammetric Computer Vision Lab\\
The Ohio State University\\
{\tt\small medicharla.2@buckeyemail.osu.edu}
\and
Alper Yilmaz\\
Photogrammetric Computer Vision Lab\\
The Ohio State University\\
{\tt\small yilmaz.15@osu.edu}
}

\begin{document}
\maketitle
\begin{abstract}
In this paper, we introduce MotivNet, a generalizable facial emotion recognition model for robust real-world application. Current state-of-the-art FER models tend to have weak generalization when tested on diverse data, leading to deteriorated performance in the real world and hindering FER as a research domain. Though researchers have proposed complex architectures to address this generalization issue, they require training cross-domain to obtain generalizable results, which is inherently contradictory for real-world application. Our model, MotivNet, achieves competitive performance across datasets without cross-domain training by using Meta-Sapiens as a backbone. Sapiens is a human vision foundational model with state-of-the-art generalization in the real world through large-scale pretraining of a Masked Autoencoder. We propose MotivNet as an additional downstream task for Sapiens and define three criteria to evaluate MotivNet's viability as a Sapiens task: benchmark performance, model similarity, and data similarity. Throughout this paper, we describe the components of MotivNet, our training approach, and our results showing MotivNet is generalizable across domains. We demonstrate that MotivNet can be benchmarked against existing SOTA models and meets the listed criteria, validating MotivNet as a Sapiens downstream task, and making FER more incentivizing for in-the-wild application. The code is available at
\href{https://github.com/OSUPCVLab/EmotionFromFaceImages}{github.com/OSUPCVLab/EmotionFromFaceImages}
\end{abstract}    
\section{Introduction}
\label{sec:intro}

As humans, we are inherently emotive and express ourselves through language and pose, using tools such as gestures, body postures, and facial expressions to visually enhance our message. Experts across various fields have recognized the importance of these visual cues, prompting researchers to analyze them systematically. One such research domain that has formed as a result is Facial Expression Recognition (FER) \cite{wang2024surveyfacialexpressionrecognition, s24113484}. FER is a computer vision discipline focused on developing models that categorize static facial images as universally recognizable emotions. It has grown significantly in recent years and has become a key focus area in computer vision due to its widespread applicability in healthcare, education, and public safety \cite{s23167092, s24113484, KHARE2024102019, wang2024surveyfacialexpressionrecognition}. 

Current state-of-the-art FER models can classify facial images on intra-dataset samples with satisfactory performance \cite{wang2024surveyfacialexpressionrecognition}. However, these models tend to have poor generalization on diverse data, leading to severely deteriorated performance in a real-world context \cite{wang2024surveyfacialexpressionrecognition, 
xu2020investigatingbiasfairnessfacial}. This is a result of the lack of publicly available FER datasets that are large, diverse, and balanced enough to mimic the real world \cite{s24113484, fan2022combatinguncertaintyclassimbalance}. Most FER datasets today are either collected in-lab or sourced from the Internet \cite{wang2024surveyfacialexpressionrecognition}. In-lab datasets generally do not have enough samples to capture the racial and cultural variance of humans, and large-scale FER datasets collected from the Internet are heavily imbalanced against certain emotions, such as disgust or contempt, leading to limited data on the intra-class variance for these labels \cite{wang2024surveyfacialexpressionrecognition, fan2022combatinguncertaintyclassimbalance}. The use of either type of dataset causes models to have weak generalization when tested cross-domain, making FER less incentivizing for use in the real world and hindering FER as a research domain. To address this, researchers have proposed complex architectures that join data across multiple domains to improve generalization \cite{wang2024surveyfacialexpressionrecognition}; however, these architectures require training on multiple domains in order to be generalizable on them, which is inherently contradictory for real-world applicability. FER models should be generalizable on unseen data to be useful in the wild.

To address these issues, we introduce MotivNet, a transfer-learning-based facial emotion recognition model that is generalizable without cross-domain training for robust real-world application. We sample a uniform distribution of classes to comprise our dataset, minimizing the impact of class imbalance, and capture the native diversity of humans by using Meta-Sapiens \cite{khirodkar2024sapiensfoundationhumanvision} as a backbone. Sapiens is a state-of-the-art human vision foundational model with four main downstream tasks: pose estimation, body part segmentation, depth estimation, and surface normal estimation \cite{khirodkar2024sapiensfoundationhumanvision}. Built on a pretrained Masked Autoencoder \cite{he2021maskedautoencodersscalablevision}, Sapiens achieved SOTA generalization on diverse real-world human data and upholds this generalization when adapted for various tasks, incentivizing us to experiment with the model's capabilities as a feature extractor. 

In this paper, we propose MotivNet as an additional Sapiens downstream task. To evaluate our model's viability, we introduce three criteria. 1) The model architecture should not vary significantly from the Sapiens architecture; 2) The data used to fine-tune the model should be similar to the data Sapiens is trained on; 3) The model should achieve results competitive with current cross-domain FER benchmarks. Our model, MotivNet, satisfies all three criteria. MotivNet employs an encoder-decoder architecture with Sapiens as the backbone and ML-Decoder \cite{ridnik2021mldecoderscalableversatileclassification} as the classification head. It is fine-tuned on the AffectNet dataset \cite{Mollahosseini_2019}. Throughout the paper, we will discuss related work in FER, the Meta-Sapiens architecture, our methods for unbiasedly training a model, and our results showing that MotivNet is generalizable. We validate MotivNet as a Sapiens task, pushing forward FER's applicability in the real world. Our contributions are summarized below.

1. We introduce MotivNet, an FER model that is generalizable without cross-domain training and finetuned in an unbiased manner for robust application in the real world.

2. MotivNet is proposed and validated as a novel Sapiens downstream task, achieving performance competitive with current cross-domain FER models.

\section{Related Work}
\label{sec:relatedWork}

% I want to give specific examples for popular FER datasets and their characteristics, mainly their class imbalance. Then talk about the types of common FER models. And then more specifically models that try to address the generalization issues and their characteristics.

\textbf{Datasets}. A few large-scale datasets published in recent years have become the norm for training and benchmarking FER models. These include AffectNet \cite{Mollahosseini_2019}, FER-2013 \cite{FER2013}, EmotioNet \cite{EmotioNet}, and RAF-DB \cite{RAF-DB} \cite{wang2024surveyfacialexpressionrecognition}. These datasets contain, at a minimum, tens of thousands of static facial images for use in FER tasks, with AffectNet and EmotioNet superseding this with around one million images \cite{wang2024surveyfacialexpressionrecognition}. To accumulate such a large number of samples, these datasets are sourced from the Internet. They are collected with the goal of providing robust real-world context that captures the native diversity of humans. However, as previously described, these large in-the-wild FER datasets are affected by heavy class imbalance \cite{fan2022combatinguncertaintyclassimbalance}. Since the images are pulled from the Internet, it is likely that a primary source for them is social media. This brings about an uneven distribution of class frequencies, since people generally do not post pictures of themselves disgusted or fearful. Models solely trained with these datasets become affected by this class imbalance, showing a significant bias toward majority labels such as happy or neutral when tested \cite{fan2022combatinguncertaintyclassimbalance}. To counteract this imbalance, we randomly select a uniform distribution of samples from each class to comprise our dataset, thus limiting the impact of this native imbalance. 

\begin{figure}[ht]
    \centering
    \includegraphics[width=160pt]{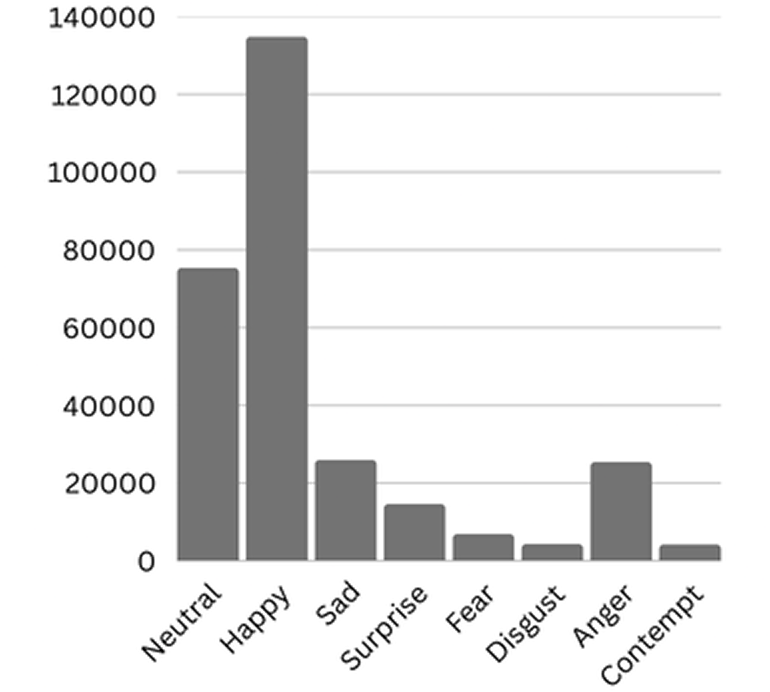}
    \caption{An example of FER class imbalance by graphing the frequency of each label in the AffectNet dataset. \cite{khirodkar2024sapiensfoundationhumanvision}}
    \label{fig:affectnetclassimbalance}
% \vspace{-0.1cm}
\end{figure}

\textbf{Common Models.} Deep Learning FER models today typically adhere to a common pipeline: a backbone that extracts features and a classification head that outputs predictions \cite{wang2024surveyfacialexpressionrecognition}. The scope of relevant FER work can be broadly categorized into two variants based on the model's backbone architecture: Convolutional Neural Networks (CNN) \cite{oshea2015introductionconvolutionalneuralnetworks} or attention-based networks (including hybrid architectures) \cite{wang2024surveyfacialexpressionrecognition, vaswani2023attentionneed}.

CNN's were the standard in FER for some time, with notable architectures such as VGGNet \cite{simonyan2015deepconvolutionalnetworkslargescale} and ResNet \cite{he2015deepresiduallearningimage} acting as the backbone. They have had remarkable performance due to CNN's ability to extract local features from images \cite{wang2024surveyfacialexpressionrecognition}. Popular work in FER used this capability to train the model on feature-rich regions in the face that directly impact emotion. However, CNNs tend to exhibit inductive biases that restrict the model, making it difficult to recognize complex global relationships \cite{jamil2022comprehensivesurveytransformerscomputer}.

Attention mechanisms in computer vision are a recent development that have made significant progress. In 2020, Google published ViT, an architecture that treats image data as a sequence of non-overlapping patches for Transformer use \cite{dosovitskiy2021imageworth16x16words, vaswani2023attentionneed}. The ViT architecture has been adopted in various domains, demonstrating excellent results because of its ability to capture long-range dependencies in image data \cite{wang2024surveyfacialexpressionrecognition}. ViT lacks the inductive biases exhibited by CNNs and is resistant to visual distortions in input data \cite{jamil2022comprehensivesurveytransformerscomputer}, making ViT a desirable backbone given sufficient training data.

\textbf{Generalization.} A few models published in recent years have attempted to address the cross-domain generalization issue by following either a transfer-learning approach or an adaptation-learning approach \cite{wang2024surveyfacialexpressionrecognition}. These include Cross-domain Sample Relationship Learning (CSRL) \cite{CSRL}, Adversarial Graph Representation Adaptation (AGRA) \cite{AGRA}, and Emotional Conditional Adaptation Network (ECAN) \cite{ECAN} \cite{wang2024surveyfacialexpressionrecognition}. CSRL, a transfer-learning model, developed both intra- and inter-transformers to understand domain-specific features and extract the similarities between domains, respectively \cite{wang2024surveyfacialexpressionrecognition, CSRL}. They implemented an alignment strategy on top to combine both results, effectively learning domain-agnostic features. \cite{wang2024surveyfacialexpressionrecognition, CSRL}. AGRA and ECAN are adaptation-learning models that learn universal features by aligning the feature distributions for each emotion class across domains \cite{AGRA, ECAN}. AGRA utilizes adversarial learning and graph propagation approaches to correlate global and local features, while ECAN focuses on statistical distribution alignment and class-specific reweighting to learn domain-invariant discriminative features \cite{AGRA, ECAN}.

\begin{table}[ht]
\renewcommand\arraystretch{1.4} % Increase row height
\setlength{\tabcolsep}{0.4pt} % Adjust the spacing between columns
\scalebox{0.85}{ % Adjust the scale to fit the single-column width
    \centering
    \begin{tabular}{
        >{\raggedright\arraybackslash}p{1.8cm} 
        >{\centering\arraybackslash}p{0.7cm} 
        >{\centering\arraybackslash}p{1.5cm} 
        >{\centering\arraybackslash}p{1.6cm} 
        >{\centering\arraybackslash}p{1.1cm} 
        >{\centering\arraybackslash}p{1.1cm} 
        >{\centering\arraybackslash}p{1.2cm}} % Removed last column
        \toprule
        \multirow{2}{*}{\textbf{Method}} & \multirow{2}{*}{\textbf{Year}} & \multirow{2}{*}{\textbf{Backbone}} & \multirow{2}{*}{\makecell{\textbf{Source} \\ \textbf{Dataset}}} & \multicolumn{3}{c}{\textbf{Target Dataset WAR}} \\ \cline{5-7}
        \multicolumn{4}{c}{} & \textbf{JAFFE} & \textbf{CK+} & \textbf{FER-2013} \\
        \midrule
        ECAN~\cite{ECAN} & 2022 & ResNet50 & RAF-DB & 57.28 & 79.77 & 56.46 \\
        AGRA~\cite{AGRA} & 2022 & ResNet50 & RAF-DB & 61.50 & 85.27 & 58.95 \\
        CSRL~\cite{CSRL} & 2023 & ResNet18 & RAF-DB & 66.67 & 88.37 & 55.53 \\
        \bottomrule
    \end{tabular}
}
\label{tab:crossDomainRelatedWork}
\caption{Performance of cross-domain FER models measured with WAR. JAFFE \cite{JAFFE} consists of in-lab collected static facial images and CK+ \cite{CK+} consists of in-lab collected videos, however, the last frame of the videos is often used for benchmarking image models. FER-2013 \cite{FER2013} consists of images collected from the wild. Adapted from \textit{A Survey on Facial Expression Recognition of Static and Dynamic Emotions (2024)} \cite{wang2024surveyfacialexpressionrecognition}.}
% \vspace{-0.5cm}
\end{table}

These models, though they have satisfactory results cross-domain, have a caveat: they require training on multiple domains in order to be generalizable on them. This caveat is inherently contradictory with real-world applicability since FER models should maintain generalization on unseen domains without training on them. Our model, MotivNet, is able to maintain generalization without cross-domain training or complex architectures, largely due to Sapiens' pretraining effort. 

\textbf{Sapiens.} Sapiens achieved SOTA generalization on their downstream tasks by first conducting a large-scale pretraining of a Masked Autoencoder (MAE), a model that learns to provide rich feature spaces for images \cite{khirodkar2024sapiensfoundationhumanvision, he2021maskedautoencodersscalablevision}. Meta pretrained their MAE with Humans-300M, a proprietary dataset consisting of approximately one billion in-the-wild human images \cite{khirodkar2024sapiensfoundationhumanvision}. The dataset is preprocessed to include only unobstructed images and cropped to the human figure \cite{khirodkar2024sapiensfoundationhumanvision}. After pretraining on around 300 million preprocessed images, their MAE was able to reconstruct the original data from minimal input signals with excellent performance. They then followed a transfer-learning approach for each downstream task by transferring their MAE into another encoder-decoder model and finetuning with task-specific data \cite{khirodkar2024sapiensfoundationhumanvision}.

\begin{figure}[ht]
    \centering
    \includegraphics[width=\linewidth]{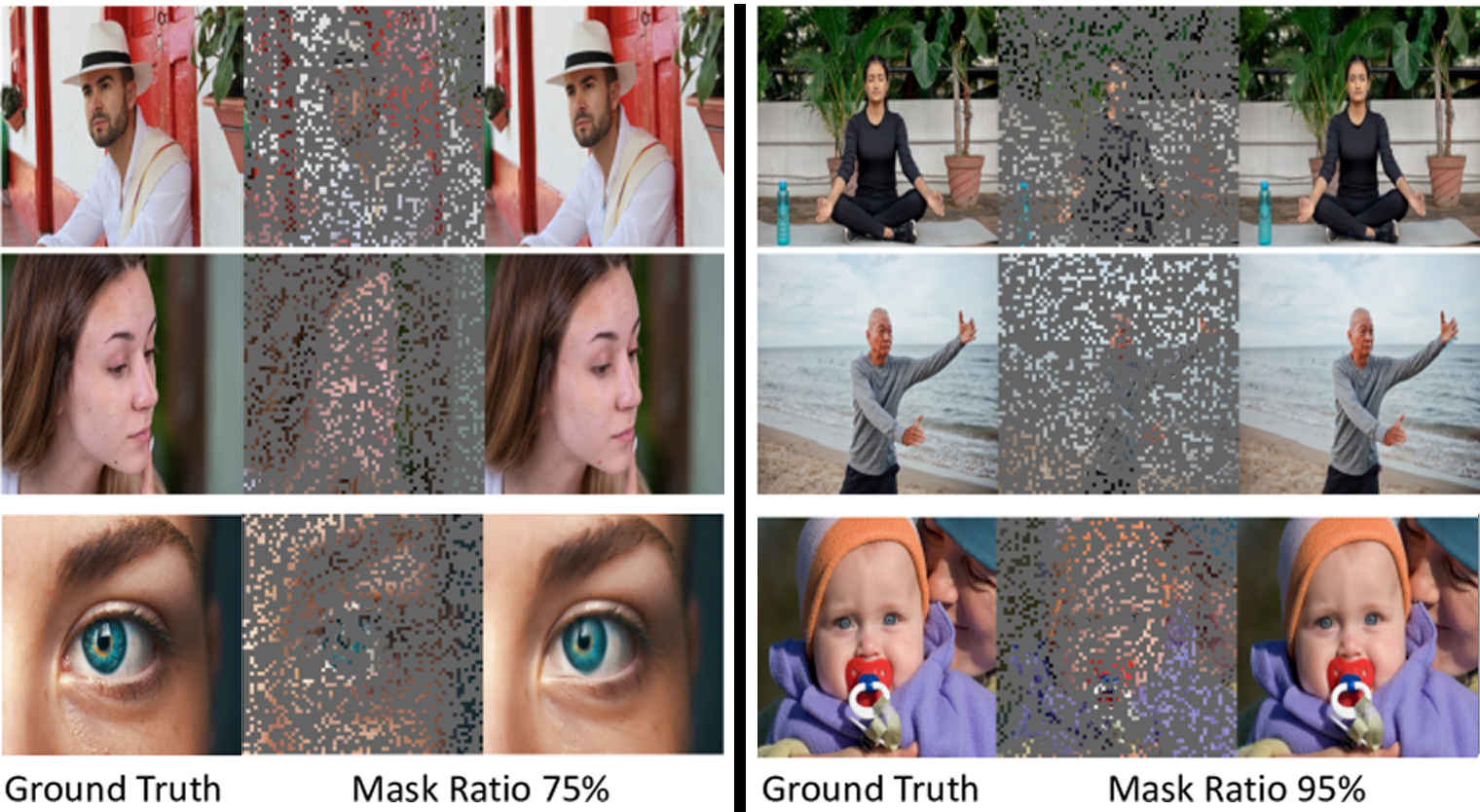}
    \caption{Sapiens MAE reconstruction. Each triplet shows ground truth on the left, the mask in the middle, and the reconstruction on the right. The bottom shows different triplets under a 75\% and 95\% mask ratio. Adapted from \textit{Sapiens: Foundation for Human Vision Models} \cite{khirodkar2024sapiensfoundationhumanvision}}
    \label{fig:sapiens}
% \vspace{-0.5cm}
\end{figure}

\section{Methods}
\label{sec:methods}

To build MotivNet we split our development into three stages described below: data selection, model architecture, and experimentation.  

\subsection{Data Selection}

We chose to train MotivNet on AffectNet for two reasons. Since AffectNet is a diverse dataset consisting of more than a million facial images, it encapsulates the racial and cultural variance of humans, enhancing MotivNet's generalization cross-domain \cite{Mollahosseini_2019}. AffectNet is also taken from the wild and shares this characteristic with Humans-300M, thus aligning MotivNet with Sapiens' original direction for robustness on in-the-wild human data and satisfying the data similarity criteria \cite{Mollahosseini_2019, khirodkar2024sapiensfoundationhumanvision}. AffectNet is manually annotated with one of eight recognizable emotions: neutral, happy, sad, surprise, fear, disgust, anger, and contempt \cite{Mollahosseini_2019}. 

\begin{figure}[ht]
    \centering
    \includegraphics[width=180pt]{imgs/affectnet.png}
    \label{fig:affectnet}
    \caption{Examples of the eight types of labeled facial images in AffectNet. Adapted from \textit{AffectNet: A Database for Facial Expression, Valence, and Arousal Computing in the Wild} \cite{Mollahosseini_2019}.}
% \vspace{-0.3cm}
\end{figure}

We chose to exclude contempt from our dataset in order to have uniformity in class labels since other popular FER datasets such as FER-2013 and JAFFE only include the seven universal emotions \cite{FER2013, JAFFE}. As mentioned, AffectNet is heavily class imbalanced since the images were sourced from the web, likely social media sources \cite{Mollahosseini_2019}. To counteract this imbalance, and train MotivNet in an unbiased manner, we sampled \textit{n} samples from each emotion class using SRS without replacement. \textit{C} is the number of classes and \textit{AN} is the dataset grouped by class.
% \vspace{-0.3cm}
\begin{align*}
    n &= \min_{c \in C} |AN_c|
\end{align*}

We sampled 3,803 images per class for a total of 26,621 static facial images for the model to be fine-tuned on. Since Sapiens' encoder has already been pretrained, there is a sufficient number of samples to fine-tune it for facial emotion recognition. AffectNet came with a presampled validation set of 500 images per class, which we use as the test set for model evaluation. The sampled set defined above is randomly split into the training and validation sets at run-time with an 8:10 and 2:10 ratio per class, respectively.

\subsection{Model}

To satisfy the architecture similarity criteria, we followed Sapiens' approach in defining models for downstream tasks: transfer a pre-trained encoder into another encoder-decoder model \cite{khirodkar2024sapiensfoundationhumanvision}.

\textbf{Encoder.} We chose to use the Sapiens 1B-parameter 308 keypoint pose estimation model \cite{khirodkar2024sapiensfoundationhumanvision} as the basis of MotivNet. It has already been finetuned to identify 274 facial keypoints out of 308 total keypoints \cite{khirodkar2024sapiensfoundationhumanvision}, so we felt this was a sufficient starting point to provide rich feature spaces on facial images.

\textbf{Decoder.} After initial experimentation, we found that attention-based classification heads had superior performance to simple multilayer perceptron classification heads. This was due to Attention's weighted sum property \cite{vaswani2023attentionneed}, which effectively enables dynamic feature selection in the feature space based on perceived importance. In practice, we chose ML-Decoder, a lightweight attention-based classification head with better spatial data utilization than global average pooling, a common accumulation algorithm \cite{ridnik2021mldecoderscalableversatileclassification, lin2014networknetwork}. ML-Decoder is a variant of the transformer-decoder architecture with the key differences being the elimination of self-attention, a group decoding scheme, and non-learnable queries \cite{ridnik2021mldecoderscalableversatileclassification, vaswani2023attentionneed}. 

\begin{figure}[ht]
    \centering
    \includegraphics[width=180pt]{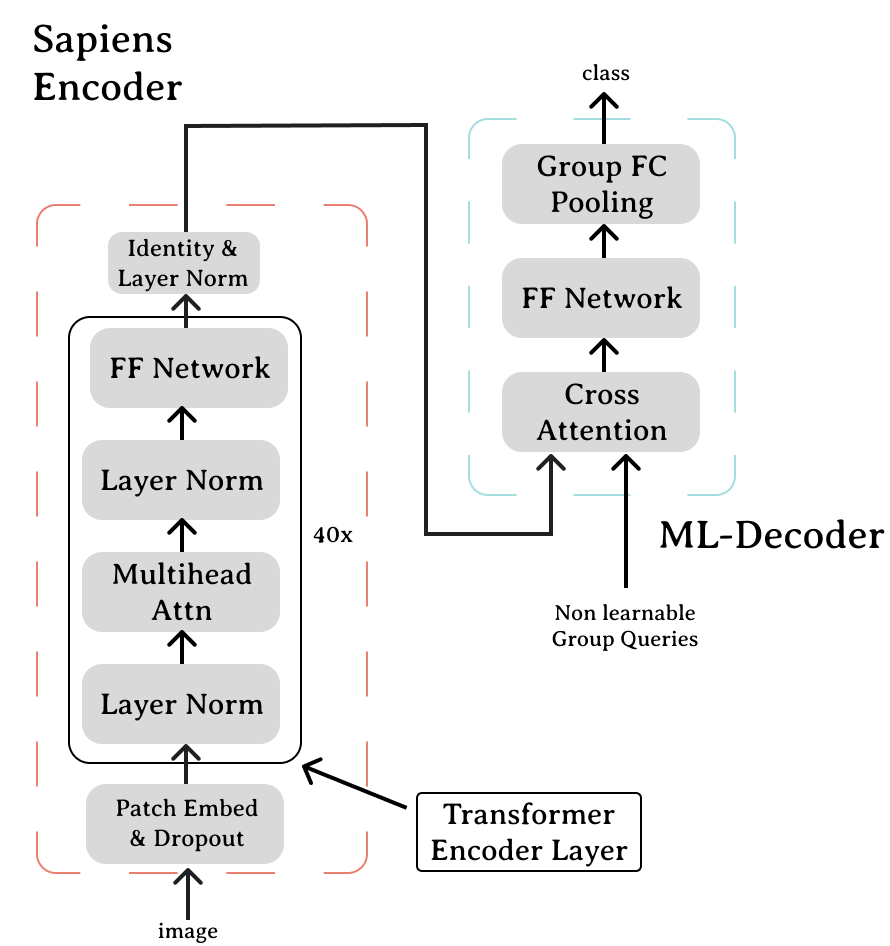}
    \caption{A diagram of MotivNet's encoder-decoder architecture. The encoder is Sapiens' network consisting of a patch embedding, dropout, and 40 transformer-encoder layers. The decoder is ML-Decoder's network consisting of a cross attention mechanism with non-learnable group queries feeding into a fully connected network and then a novel group fully connected pooling. Adapted from \textit{Sapiens: Foundation for Human Vision Models} and \textit{ML-Decoder: Scalable and Versatile Classification Head}  \cite{khirodkar2024sapiensfoundationhumanvision,ridnik2021mldecoderscalableversatileclassification}.}
    \label{fig:architecture}
% \vspace{-0.3cm}
\end{figure}

\subsection{Experimentation}

Our implementation includes training a baseline model for forty epochs with a seed of forty-two for random initialization. We trained the baseline on three NVIDIA RTX A6000 with the AdamW optimizer \cite{loshchilov2017sgdrstochasticgradientdescent} and the Cosine Annealing with Warm Restarts learning rate scheduler \cite{loshchilov2019decoupledweightdecayregularization}. This resulted in stable weight updates without excessive regularization and randomly assigned learning rates, thereby enhancing generalization and improving convergence on optimal solutions. This optimizer and scheduler were also used by Sapiens in their downstream tasks, further enhancing architectural similarity \cite{khirodkar2024sapiensfoundationhumanvision}.

Since the encoder is being finetuned while the decoder is being trained from scratch, we set different learning rates for each. The encoder is being finetuned with an initial learning rate of 1E-7, while the decoder is being trained with an initial learning rate of 1E-5. 

MotivNet was trained with a batch size of two and CrossEntropy as the loss function, the standard for multi-label classification tasks. We used a gradient accumulation protocol for every eight batches for stable gradient updates, effectively setting the batch size to sixteen. Before training, each image is interpolated from (224,224) pixel dimensions to (768,768) pixel dimensions and padded vertically with white space for a final dimension of (768,1024). The images are also normalized across each channel, as was done with the Sapiens 308 keypoint estimator \cite{khirodkar2024sapiensfoundationhumanvision}.

\section{Results}
\label{sec:results}

Once trained, we reviewed MotivNet's training performance. MotivNet reached convergence within the first thirty epochs, largely due to Meta's pretraining effort. With limited task-specific data, MotivNet achieved its best performance at epoch twenty-seven.

MotivNet achieved weighted average recall (WAR) scores that make it competitive with current cross-domain FER models and top-2 accuracy scores that are comparable to state-of-the-art domain-specific FER models. Results are shown below.

\begin{table}[ht]
\renewcommand\arraystretch{1.4} % Matches the 1.4 stretch of your first table
\setlength{\tabcolsep}{0.4pt} % Adjust the spacing between columns      % Standard spacing for better readability
\centering
\scalebox{0.85}{ % Scale to match the style of your main results table
    \begin{tabular}{
        >{\raggedright\arraybackslash}p{2.2cm} 
        >{\centering\arraybackslash}p{2.0cm} 
        >{\centering\arraybackslash}p{2.0cm} 
        >{\centering\arraybackslash}p{1.9cm}
        >{\centering\arraybackslash}p{1.9cm}}
        \toprule
        \textbf{Dataset} & \textbf{WAR} & \textbf{Top-2 Acc} & \textbf{Precision} & \textbf{F1} \\
        \midrule
        JAFFE \cite{JAFFE}                 & 58.57 & 76.19 & 75.25 & 56.20\\
        CK+ \cite{CK+}                     & 80.00 & 96.67 & 70.77 & 76.15\\
        FER-2013 \cite{FER2013}             & 53.87 & 74.80 & 49.59 & 49.13\\
        AffectNet \cite{Mollahosseini_2019} & 62.52 & 83.50 & 62.63 & 62.41\\
        \bottomrule
    \end{tabular}
}
\caption{MotivNet's results benchmarked across the JAFFE \cite{JAFFE}, CK+ \cite{CK+}, FER-2013 \cite{FER2013} and AffectNet \cite{Mollahosseini_2019} FER datasets using Weighted Average Recall (WAR), Top-2 Accuracy, Precision, and F1 metrics. The results are collected using only the seven emotion labels MotivNet was trained on. }
\label{tab:results_p1} 
\end{table}

\begin{table}[ht]
\renewcommand\arraystretch{1.4} % Increase row height
\setlength{\tabcolsep}{0.4pt} % Adjust the spacing between columns
\scalebox{0.85}{ % Adjust the scale to fit the single-column width
    \centering
    \begin{tabular}{
        >{\raggedright\arraybackslash}p{2.2cm} 
        >{\centering\arraybackslash}p{2.2cm} 
        >{\centering\arraybackslash}p{1.2cm} 
        >{\centering\arraybackslash}p{1.2cm} 
        >{\centering\arraybackslash}p{1.6cm}
        >{\centering\arraybackslash}p{1.6cm}} 
        \toprule
        \multirow{2}{*}{\textbf{Method}} & \multirow{2}{*}{\textbf{Backbone}} & \multicolumn{4}{c}{\textbf{Target Dataset WAR}} \\ \cline{3-6}
        \multicolumn{2}{c}{} & \textbf{JAFFE} & \textbf{CK+} & \textbf{FER-2013} & \textbf{AffectNet} \\
        \midrule
        ECAN~\cite{ECAN} & ResNet50 & 57.28 & 79.77 & 56.46 & 51.84 \\
        AGRA~\cite{AGRA} & ResNet50 & 61.50 & 85.27 & 58.95 & -- \\
        CSRL~\cite{CSRL} & ResNet18 & 66.67 & 88.37 & 55.53 & -- \\
        \bottomrule
        \textbf{MotivNet(Ours)} & \textbf{ViT} & \textbf{58.57} & \textbf{80.00} & \textbf{53.87} & \textbf{62.52} \\
        \bottomrule
    \end{tabular}
}
\caption{Performance comparison of MotivNet against other cross-domain FER models. Results are reported using Weighted Average Recall across both lab-controlled (JAFFE \cite{JAFFE}, CK+ \cite{CK+}) and in-the-wild (FER-2013 \cite{FER2013}, AffectNet \cite{Mollahosseini_2019}) datasets. Adapted from \textit{A Survey on Facial Expression Recognition of Static and Dynamic Emotions (2024)} \cite{wang2024surveyfacialexpressionrecognition}.}
% \vspace{-0.5cm}
\label{tab:results_p2}
\end{table}

\begin{table}[ht]
\renewcommand\arraystretch{1.4} % Matches the 1.4 stretch of your first table
\setlength{\tabcolsep}{0.4pt} % Adjust the spacing between columns      % Standard spacing for better readability
\centering
\scalebox{0.85}{ % Scale to match the style of your main results table
    \begin{tabular}{
        >{\raggedright\arraybackslash}p{2.2cm} 
        >{\centering\arraybackslash}p{3cm} 
        >{\centering\arraybackslash}p{1.7cm}
        >{\centering\arraybackslash}p{1.7cm}
        >{\centering\arraybackslash}p{1.7cm}}
        \toprule
        \textbf{Dataset} & \textbf{SOTA Model} & \textbf{SOTA Top-1 Acc} & \textbf{MotivNet Top-1 Acc} & \textbf{MotivNet Top-2 Acc}\\
        \midrule
        JAFFE \cite{JAFFE} & DenseNet-161 \cite{roy2021facial} & 99.52 & 58.57 & 76.19\\
        CK+ \cite{CK+} & PAtt-Lite \cite{Ngwe_2024} & 100 & 80.00 & 96.66\\
        FER-2013 \cite{FER2013} & EfficientFER \cite{11017006} & 82.47 & 53.87 & 74.80\\
        AffectNet \cite{Mollahosseini_2019} & ResEmoteNet \cite{roy2024resemotenetbridgingaccuracyloss} & 72.93 & 62.52 & 83.50\\
        \bottomrule
    \end{tabular}
}
\caption{Performance comparison of MotivNet against the SOTA domain-specific FER models. Results are reported using the SOTA Top-1 Accuracy scores and MotivNet's Top-1 and Top-2 Accuracy scores on seven emotion labels.}
\label{tab:results_p3}
\end{table}

Through these results, we show that MotivNet fulfills the final criteria for benchmark performance. As displayed in Table \ref{tab:results_p2}, MotivNet achieved Weighted Average Recall (WAR) scores competitive with established cross-domain FER architectures across diverse datasets. MotivNet was able to generalize facial emotion feature extraction without complex architectures or cross-domain training. In Table \ref{tab:results_p3}, MotivNet was compared against SOTA models on each dataset that were optimized for single-domain environments. MotivNet yielded Top-2 accuracy scores within ten percent of the SOTA models on the CK+ \cite{CK+} and FER-2013 \cite{FER2013} datasets, and a Top-1 accuracy score within ten percent of the SOTA model on AffectNet \cite{Mollahosseini_2019}.
\section{Conclusion}
\label{sec:conclusion}

In this paper, we present MotivNet, a generalizable model for FER tasks designed to bridge the gap between constrained benchmark datasets and real-world application. While established cross-domain FER models rely on intricate architectures or cross-domain training to achieve generalization, MotivNet leverages Meta's Sapiens \cite{khirodkar2024sapiensfoundationhumanvision} model to eliminate any specialized cross-domain optimization, and yielding performance competitive with existing cross-domain architectures. By framing MotivNet as a Sapiens downstream task, we introduced a set of robust evaluation criteria that both guided our development processes and validated the applicability of Sapiens across diverse environments. Our findings suggest that a viable path towards real-world FER lies in the large-scale pretraining of human-centric foundational models and can be explored further to apply FER in unconstrained real-world environments.

\end{document}